\documentclass{article}

\usepackage{iclr2027_conference,times}
\iclrfinalcopy

\usepackage[utf8]{inputenc}
\usepackage[T1]{fontenc}
\usepackage{amsmath,amssymb,amsfonts}
\usepackage{booktabs}
\usepackage{array}
\usepackage{algorithm}
\usepackage{algpseudocode}
\usepackage{graphicx}
\usepackage{url}
\usepackage[table]{xcolor}
\usepackage{arydshln}
\usepackage{microtype}
\usepackage{multirow}
\usepackage{caption}
\usepackage{subcaption}
\usepackage{float}
\usepackage{ifthen}
\usepackage{placeins}
\usepackage[breaklinks=true,colorlinks=true,citecolor=blue,linkcolor=blue,urlcolor=blue]{hyperref}

\newcommand{\Para}{\mathrm{Para}}

\newcommand{\margin}{m}

\newcommand{\figplaceholder}[2]{%
  \fbox{%
    \begin{minipage}[c][#1][c]{0.96\linewidth}%
      \centering\small #2%
    \end{minipage}%
  }%
}

\newsavebox{\saetablebox}
\newcommand{\fittablefontto}[3]{%
\begingroup
  #2%
  \sbox{\saetablebox}{#3}%
  \ifdim\wd\saetablebox>#1%
    \resizebox{#1}{!}{\usebox{\saetablebox}}%
  \else%
    \usebox{\saetablebox}%
  \fi%
\endgroup
}
\newcommand{\fittableto}[2]{\fittablefontto{#1}{\small}{#2}}
\newcommand{\fittable}[1]{\fittableto{\textwidth}{#1}}

\title{Active Budget Can Kill Sensitivity:\\ Diagnosing and Repairing TopK Sparse Autoencoder Reliability}

\author{%
Zhenting Huang$^{1}$ \qquad Bo Jiang$^{1,*}$ \qquad Junnan Liu$^{3}$ \qquad Zhixing Tan$^{2}$  \qquad Qianren Mao$^{2,*}$ \\
\small $^{1}$Beihang University \qquad $^{2}$Zhongguancun Laboratory \qquad $^{3}$Monash University\\
\small \texttt{trend@buaa.edu.cn, jiangbo@buaa.edu.cn, maoqr@zgclab.edu.cn}%
}

\begin{document}
\maketitle

\fancyhead{}
\lhead{Preprint. Under review.}

\begin{abstract}
Sparse autoencoders (SAEs) are increasingly scaled to wider dictionaries to recover fine-grained structure from large language model activations. However, a feature is useful for interpretation only if it remains a stable unit of analysis when the same meaning is expressed in different surface forms. We study this reliability question for TopK SAEs through \textit{feature sensitivity}: the probability that a source-active feature remains active under meaning-preserving paraphrases. Experiments demonstrate that practical scaling selectively reduces the sensitivity of rare features, while common features remain comparatively stable. A controlled width$\times k$ factorial locates the cause in the active budget $k$: rare-feature sensitivity declines monotonically as $k$ grows while reconstruction error improves over the same range, so the loss is a property of the selection boundary rather than of width alone. We trace this failure to the geometry of TopK selection: rare active features often lie close to the cutoff between selected and rejected features, so small paraphrase-induced shifts can reorder nearby competitors and remove them from the active set. The distance to that cutoff, the \textbf{active margin}, predicts which features are lost without any threshold, consistently across depths and model families. Guided by the margin diagnosis, we introduce \textbf{pairwise rank stabilization}, which targets the source--paraphrase ordering failure at the cutoff and raises rare-feature sensitivity by $8.83$ percentage points on sources that played no role in selecting the objective or its hyperparameters, while reconstruction and alive-feature coverage stay close to the baseline. Overall, our results suggest that wide TopK SAEs should be evaluated not only by reconstruction, sparsity, and feature count, but also by feature reliability under semantic variation and the boundary geometry that determines whether features remain available as stable units of analysis.
\end{abstract}

\section{Introduction}

Large language models (LLMs) have demonstrated remarkable capabilities across reasoning, coding, and open-ended generation, yet their internal mechanisms remain difficult to inspect. This opacity limits our ability to understand when models rely on robust abstractions, spurious correlations, or brittle heuristics. Mechanistic interpretability addresses this challenge by translating neural network behavior into inspectable computational structures, including transformer circuits, causal tracing, and feature-based explanations~\citep{elhage2021framework,wang2023interpretability,meng2022locating,openai2023autointerp}. For such analyses to be reliable, however, the units of explanation must be not only interpretable, but also stable across semantically equivalent inputs.

Sparse autoencoders (SAEs) have emerged as a central tool for extracting such units from LLM activations. Rather than analyzing polysemantic neurons directly, SAEs learn overcomplete sparse dictionaries whose latent features can be interpreted through activating examples, circuit roles, and causal interventions~\citep{bricken2023monosemanticity,cunningham2023sparse,templeton2024scaling,gao2024scaling}. Scaling SAE width is especially appealing because larger dictionaries can recover finer-grained structure that narrower models may merge or miss. Yet this benefit raises a reliability question: the additional fine-grained features exposed by wider SAEs may look coherent on their top-activating examples and still disappear when the same meaning is expressed in a different surface form. Recent SAE and circuit-basis evaluations sharpen this concern: principled task evaluations, synthetic benchmarks, canonical-unit critiques, and random-baseline checks show that reconstruction can overstate latent quality, while neuron-basis circuit results caution that learned sparse dictionaries are not the only possible units of analysis~\citep{makelov2025principled,chanin2026synthsaebench,leask2025canonical,korznikov2026sanitychecks,arora2026neuronbasis}.

\begin{figure*}[t]
  \centering
  \IfFileExists{figures/overall-01.pdf}{\includegraphics[width=0.98\textwidth]{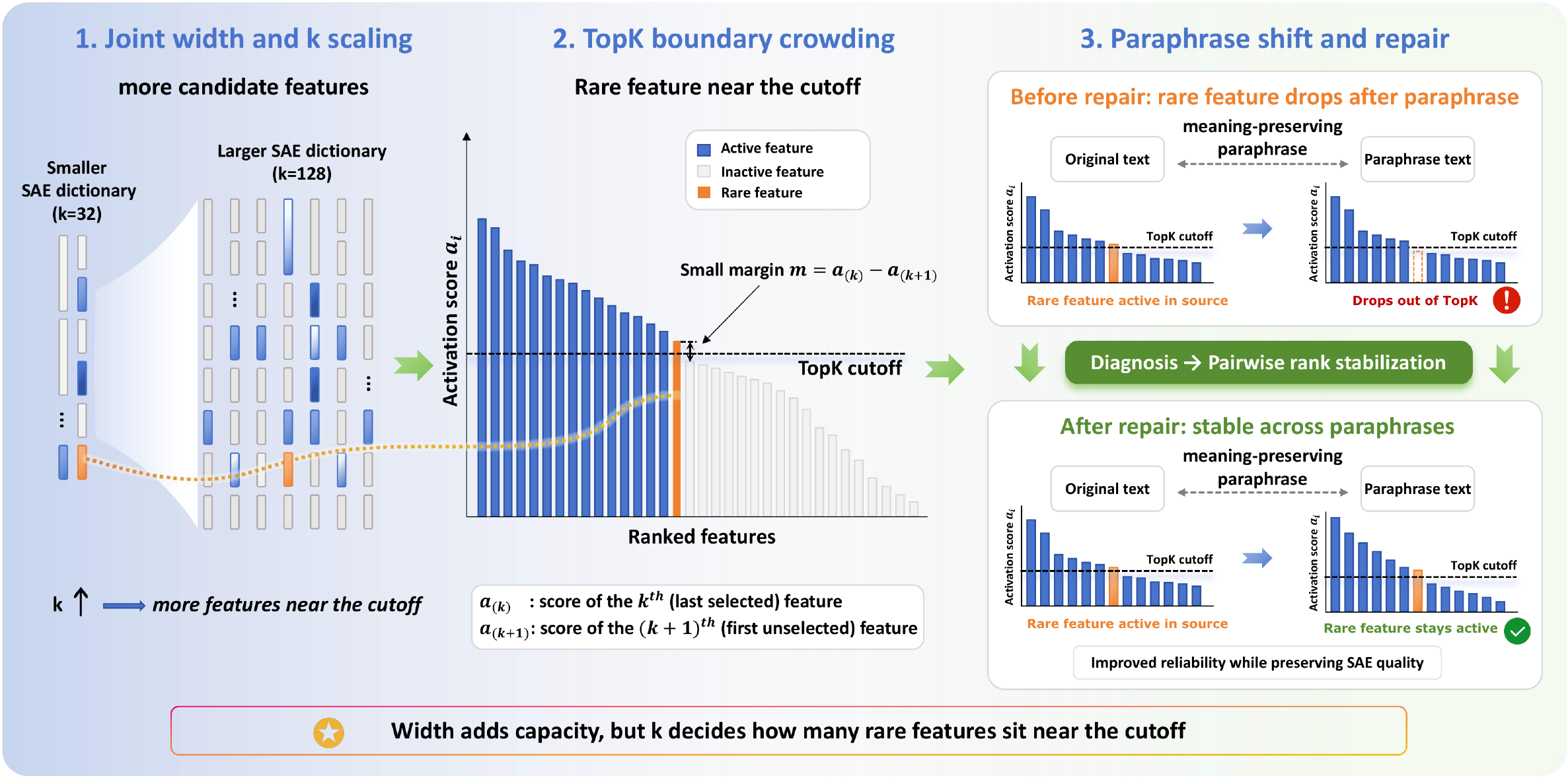}}{\IfFileExists{figures/overall-01.png}{\includegraphics[width=0.98\textwidth]{figures/overall-01.png}}{\figplaceholder{1.10in}{Figure 1 placeholder: overall mechanism illustration.}}}
  \caption{\textbf{Mechanism overview.} Joint scaling grows dictionary width and the active budget together, reveals rare features that look coherent on their top activating examples, and moves those features toward the selection boundary, where a paraphrase-induced shift can reorder the active set.}
  \label{fig:overview}
\end{figure*}

We study this reliability gap through \textbf{feature sensitivity}: the probability that a feature active on a source input remains active under meaning-preserving paraphrases~\citep{tian2025sensitivity}. Feature sensitivity operationalizes a basic desideratum for SAE-based interpretation: if an explanation, audit, or intervention relies on a feature, that feature should remain available when the underlying meaning is preserved. Paraphrases provide a natural probe because they retain the intended semantic content while varying wording, tokenization, and the local residual-stream geometry.

Our central hypothesis is that the reliability of rare features depends on how crowded the TopK selection boundary is, and that practical SAE scaling crowds it. Wide SAEs are valuable precisely because they can expose specialized, low-frequency structure, but these rare features may face stronger competition for limited active slots under TopK selection. To test this hypothesis, we group features by activation frequency and evaluate their sensitivity under source-disjoint paraphrase evaluation. In GPT-2 layer 8 TopK SAEs with per-token active budgets, rare-feature sensitivity falls along the practical scaling path ($0.650$ to $0.585$), while common features remain highly stable.

The failure is selective, and a controlled factorial identifies the active budget as the factor that carries it: the width marginal means are nearly flat, and reconstruction error improves over the same range, so reliability moves in the opposite direction from fit. We trace the loss to the geometry of TopK selection, where rare active features lie closer to the cutoff and thin margins make them easier to dislodge under paraphrase-induced shifts.

Based on this diagnosis, we propose a targeted repair for rare-feature instability. Rather than globally thickening the TopK boundary, which can harm reconstruction quality and alive feature coverage, we introduce a pairwise rank stabilization objective that directly targets the observed ordering failure: features active on source inputs should stay ranked above their hard paraphrase competitors, which stabilizes the active set where failures are most likely. We compare it against externally motivated paraphrase-invariance objectives as well as boundary-aware ones. Figure~\ref{fig:overview} summarizes the chain we trace, and our results suggest that reliable SAE scaling requires not only recovering more features but also ensuring that the ones unlocked by width stay stable under semantically preserving variation.

\textbf{Contributions.} We identify a selective reliability failure of rare features in TopK SAEs and trace it to the active budget through a controlled factorial under source-disjoint paraphrase evaluation. We give a threshold-free margin diagnosis that links thin margins to active-set flips across GPT-2 depths and in Qwen and Gemma checks. We introduce pairwise rank stabilization, which raises rare-feature sensitivity by $+8.83$pp on held-out sources while preserving reconstruction quality and alive coverage.

\section{Related Work}

\paragraph{Units of Analysis in Mechanistic Interpretability.}
Mechanistic interpretability seeks units that can support explanation, circuit localization, and intervention~\citep{olah2020zoom,elhage2021framework,wang2023interpretability,conmy2023acdc,nanda2023othello,meng2022locating}; probing, attribution, and concept-based explanations pursue the same goal with internal directions or components~\citep{bau2017network,kim2018tcav,belinkov2022probing,ribeiro2016lime,sundararajan2017axiomatic,lundberg2017unified}. Yet the choice of unit is itself a scientific assumption: transformer circuits and model editing show that localized components can mediate factual behavior~\citep{geva2021transformer,geva2022transformer,bau2020understanding}, while neuron-basis circuit tracing cautions that learned sparse bases are not always the only faithful basis for analysis~\citep{arora2026neuronbasis}. This makes reliability a property of the unit itself, not merely of the explanation built around it: an explanation or intervention is difficult to reproduce if the underlying feature disappears when the same meaning is expressed differently. We therefore follow the SAE line while treating stability as part of usability. A latent should not only admit an interpretation in isolation, but also remain available when the input meaning is preserved.

\paragraph{Sparse Autoencoders and Scaling.}
SAEs adapt sparse coding and dictionary learning to LLM activations, where superposition motivates learning overcomplete sparse dictionaries~\citep{olshausen1997sparse,lee2006efficient,aharon2006ksvd,mairal2009online,makhzani2013ksparse,elhage2022superposition,cunningham2023sparse,bricken2023monosemanticity}. Scaling this recipe is appealing because wider dictionaries can improve reconstruction and expose finer feature structure~\citep{templeton2024scaling,gao2024scaling}, which sets up the reliability question studied here: the additional low-frequency features recovered by width are useful only if they remain stable enough to support interpretation. Sparsity can be imposed by different selectors: TopK SAEs impose a per-example rank cutoff~\citep{makhzani2013ksparse,gao2024scaling}, BatchTopK shifts competition to the batch and uses an estimated threshold at inference~\citep{bussmann2024batchtopk}, and JumpReLU learns per-feature thresholds~\citep{rajamanoharan2024jumprelu}. These mechanisms differ in where competition is resolved and how the boundary of feature availability is defined. We focus on TopK SAEs because each input has an explicit per-example cutoff: only the highest-scoring $k$ features can enter the selected set, so small score changes can alter which features remain active. This makes margins and rank crossings measurable diagnostics for the active-set changes induced by paraphrasing, while tying the observed instability to a concrete selection boundary.

\paragraph{SAE Evaluation Beyond Reconstruction.}
Recent SAE evaluation work strengthens the case for reliability metrics beyond reconstruction: explanation, absorption, and consistency studies ask whether learned features remain coherent as objects of analysis~\citep{openai2023autointerp,chanin2024absorption,song2025consistency}, and principled tasks, synthetic ground truth, canonical-unit critiques, and random-baseline checks show why this cannot be inferred from low reconstruction error alone~\citep{makelov2025principled,chanin2026synthsaebench,leask2025canonical,korznikov2026sanitychecks}. SAE features are also increasingly used as functional handles for representation steering and feature-based detection~\citep{zou2023representation,patel2026uncertaintycorrectness,chensae2026hallusae}, and a handle is only as reliable as the latent behind it. Controlled input variations have long exposed shortcut features and local decision-boundary failures, with paraphrase-based adversarial examples preserving meaning while changing surface form~\citep{jia2017adversarial,mccoy2019right,ribeiro2020beyond,gardner2020evaluating,morris2020textattack,iyyer2018adversarial}, and the same pairing motivates feature-consistency pressure in SAE evaluation and contrastive alignment in representation learning~\citep{song2025consistency,chen2020simclr}. Tian et al.'s feature sensitivity makes the desideratum explicit by asking whether a feature remains active on semantically similar text~\citep{tian2025sensitivity}. 
Paraphrases serve a narrower diagnostic role here: they let us test whether the same SAE latent remains active when the underlying meaning is preserved but the wording changes. We then ask three related questions: how this reliability degrades under TopK scaling, whether a boundary-based signal can predict which features will fail, and whether that diagnosis suggests an effective repair.

\section{Preliminaries}\label{sec:preliminaries}

\subsection{Setup}

\paragraph{TopK sparse autoencoders.}
An SAE maps a language model activation $z$ to a sparse latent representation and then reconstructs $z$ from the active latents. We use \emph{SAE feature} or \emph{latent} for one coordinate of this learned sparse representation, a candidate unit of analysis rather than a guaranteed semantic concept. In the TopK SAEs studied here, the encoder assigns a score to each latent feature, keeps only the $k$ highest scoring features (the \textbf{active budget}), and applies a ReLU, so a feature is \textbf{active} exactly when its post-ReLU value is positive, $f_i>0$. Entering the top-$k$ set is necessary but not sufficient for activation: as $k$ grows the $k$th largest score is frequently negative, and it is the ReLU sign that finally zeroes a borderline feature. We treat the positive latents as the SAE features available for interpretation on that input. This fixed active set is useful for controlling sparsity, but it also creates a discrete selection boundary: a feature can stop being active not because its evidence disappears, but because nearby competitors outrank it.

\paragraph{Feature sensitivity.}
Feature sensitivity evaluates whether an SAE feature remains active under meaning preserving rewrites. A feature \textbf{drops out} when it is active on the source text and inactive on the paraphrase. In a TopK SAE a dropout has two distinct sources: the feature's score falls below the $k$th largest competitor score, so it leaves the selected set; or it stays among the top $k$ but its score is non-positive, so it is zeroed after selection. We report the two sources separately, because the first is a boundary ordering event and the second is a magnitude event. Let $A_i(x)$ denote whether feature $i$ activates on input $x$, and let $\Para(x)$ be a distribution over paraphrases of $x$. We define feature sensitivity as
\begin{equation}
S_i = \mathbb{E}_{x: A_i(x)=1}\, \mathbb{P}_{x' \sim \Para(x)}[A_i(x')=1].
\end{equation}
A highly sensitive feature remains active across paraphrases; a brittle feature may look coherent on top examples but fail under equivalent surface forms. Sensitivity is a reliability probe, not a semantic oracle: some low frequency features may intentionally encode lexical or syntactic details that a paraphrase changes. We ask whether active features remain available as stable units of analysis, and whether failures can be explained by the geometry of the selection boundary. Sensitivity is measured on the paired representation that the pair-based objectives act on.

\paragraph{Feature frequency buckets.}
A feature is \emph{alive} if its post-ReLU activation is positive on at least one activation vector; \emph{alive coverage} is the percentage of dictionary features that are alive, reported alongside every intervention so reliability gains can be read separately from changes in feature count. Activation frequency is estimated on a separate frequency-estimation partition rather than the SAE training split or the paraphrase evaluation pairs, and features are grouped into rare $(0.005,0.05)$, medium $[0.05,0.5)$, and common (at least $0.5$). Alive features below the rare lower bound count toward coverage but are excluded from the bucketed averages, where their support is too small for stable feature-conditioned estimates. The buckets are diagnostic, not semantic labels.

\paragraph{Top-$k$ margin.}
In the TopK SAEs studied here, the encoder produces scores $a(z)$ and only the highest ranked features enter the active set. We define the top-$k$ margin as
\begin{equation}
\margin(z) = a_{(k)}(z) - a_{(k+1)}(z),
\end{equation}
where $a_{(j)}(z)$ is the $j$th largest encoder score. We call examples with $\margin<0.01$ \textbf{near-cutoff} examples, and use their fraction as a simple warning signal for a crowded TopK boundary. The threshold is an operational diagnostic on each SAE's own encoder-score scale, so the cross-model checks below read the near-cutoff share within each model, alongside rank and margin-bin trends. For an already active feature, its active margin is its own score minus the top-$k$ cutoff; margin bins sort active feature instances from closest to cutoff to farthest from cutoff. We call \emph{hard paraphrase competitors} the paraphrase-side features just below the cutoff that can displace a source-active feature after rephrasing.

\paragraph{Experimental setup.}
Token budgets count residual vectors at the token level, not source documents. Activations and paraphrase pairs are partitioned by source text into disjoint subsets for training, frequency estimation, source selection, and evaluation, so no source appears twice. Two pair collections are used: a \textbf{validation collection} for model comparison and selection, and a \textbf{held-out collection} of source texts that played no role in selecting the objective or its hyperparameters, which provides the independent confirmation. Qwen and Gemma checks use \texttt{Qwen/Qwen2.5-1.5B-Instruct}~\citep{qwen2025qwen25} and \texttt{google/gemma-2-9b}~\citep{gemma2024gemma2} under the same scale, partitioning, and evaluation.

\paragraph{Statistical reporting.}
All reported means are computed over independent training seeds, with error bars denoting standard error across seeds; $n$ is the number of independently trained SAEs. Pairwise comparisons use two sided Welch tests and Cohen's $d$ over seed level bucket means when both sides have at least two completed seeds. Boundary interventions are reported together with sensitivity, near-cutoff share, reconstruction, and alive-feature coverage, so reliability gains can be read separately from quality tradeoffs.

\section{Wide TopK SAEs Can Lose Feature Sensitivity}\label{sec:challenge-diagnosis}

\subsection{Scaling exposes a sensitivity failure for rare features}

The low-frequency regime is the natural stress test for width scaling: those are the features wider dictionaries are meant to add, and they support interpretation only if they stay stable enough for audits, circuit analysis, and interventions. We therefore ask whether the studied scaling regime preserves their reliability.

On the validation collection, the 10M token GPT-2 width sweep shows rare-feature sensitivity dropping from 0.650$\pm$0.015 at width 3072 to 0.585$\pm$0.017 at width 12288. By contrast, common features remain highly stable, with sensitivity changing from 0.998 to 0.956, and an intermediate 6144/$k{=}64$ check already sits in the lower-sensitivity regime. Figure~\ref{fig:width-margin}a follows that path.

The loss is selective, falling on rare features while common features hold, and it appears at every token budget we trained: rare-feature sensitivity drops along the same scaling path at 200k ($0.802$ to $0.793$), 1M ($0.779$ to $0.748$), and 10M tokens ($0.650$ to $0.585$). In this regime, rare features become less reliable as width and the active budget grow together.

\subsection{Which factor drives the decline: the active budget}

The practical scaling path raises dictionary width and the active budget together, so we separate them with a controlled factorial on the validation collection. We run a full factorial over width $\in\{3072,6144,12288\}$ and $k\in\{32,64,128\}$, three seeds per cell, with the token budget, partitions, optimizer, schedule, and paraphrase evaluation held fixed. Table~\ref{tab:widthk-factorial} reports the marginal means.
Rare-feature sensitivity falls monotonically with the active budget, by $-14.8$pp on average from $k=32$ to $k=128$, while the width marginal means span about $1.9$pp. The near-cutoff share tracks the same factor, rising from 10\% to 27\% to 54\% as $k$ grows and staying flat across width. Reconstruction MSE moves the other way, improving from 1.35 to 0.89 at width 12288 as $k$ grows. The active budget therefore drives reconstruction accuracy and rare-feature reliability in opposite directions: a single knob improves fit while crowding the boundary, and the factorial locates that knob.

\begin{table}[!t]
\centering
\caption{\textbf{Active budget versus dictionary width on the validation collection.} Marginal means of rare-feature sensitivity over the $3\times3$ width$\times k$ factorial, three seeds per cell at the 10M token budget. \emph{Rare} is the mean rare-feature sensitivity at that level of the factor; the rightmost column is in percentage points, signed for the active budget and the level span for width.}
\label{tab:widthk-factorial}
\begingroup
\setlength{\tabcolsep}{7pt}
\renewcommand{\arraystretch}{1.05}
\fittable{%
\begin{tabular}{@{}llcr@{}}
\toprule
\bfseries
Factor & \bfseries Levels & \bfseries Rare $\uparrow$ & \bfseries Rare change (pp) \\
\midrule
\normalfont
active budget $k$ & 32 / 64 / 128 & 0.645 / 0.591 / 0.497 & $-14.8$ \\
dictionary width & 3072 / 6144 / 12288 & 0.575 / 0.589 / 0.570 & $1.9$ \\
\bottomrule
\end{tabular}}
\endgroup
\end{table}

\subsection{Thin top-$k$ margins diagnose the failure}

The selective drop suggests a geometric mechanism. In a TopK SAE, a feature is active only if it outranks enough competitors to enter the active set. Thus, even if the underlying representation is largely preserved under paraphrasing, a feature near the cutoff can be displaced by a small perturbation. We therefore test whether the severe regime produces thinner margins and whether source-side margins predict paraphrase-induced flips.

\begin{figure*}[t]
  \centering
  \begin{subfigure}[t]{0.315\textwidth}
    \centering
    \IfFileExists{figures/v2_fig2_panel_a_width_pathology.pdf}{\includegraphics[width=\linewidth]{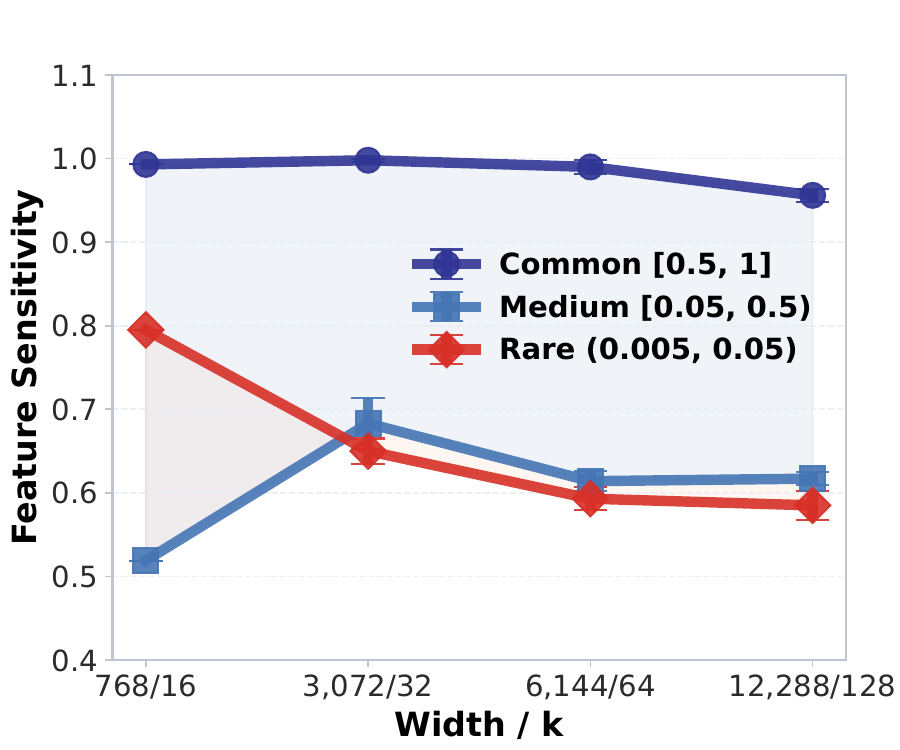}}{\figplaceholder{1.00in}{Figure 2a placeholder: width scaling curve.}}
    \caption{Width sweep: sensitivity by frequency bucket.}
  \end{subfigure}\hfill
  \begin{subfigure}[t]{0.315\textwidth}
    \centering
    \IfFileExists{figures/v2_fig2_panel_b_near_cutoff_share.pdf}{\includegraphics[width=\linewidth]{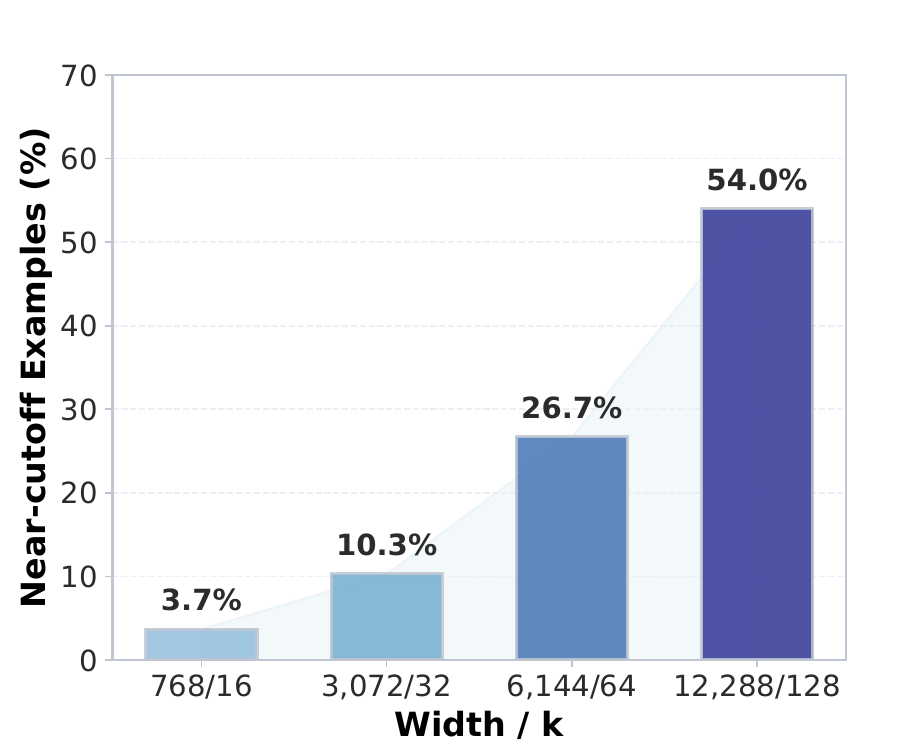}}{\figplaceholder{1.00in}{Figure 2b placeholder: near-cutoff share.}}
    \caption{Near-cutoff share ($m<0.01$).}
  \end{subfigure}\hfill
  \begin{subfigure}[t]{0.315\textwidth}
    \centering
    \IfFileExists{figures/v2_fig2_panel_c_margin_flip_evidence.pdf}{\includegraphics[width=\linewidth]{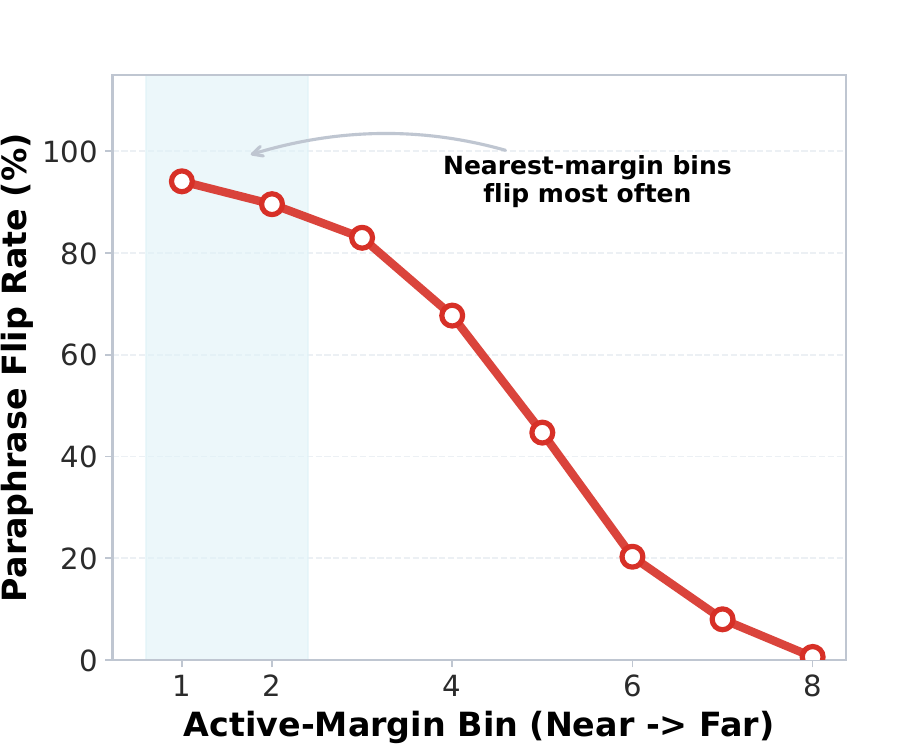}}{\figplaceholder{1.00in}{Figure 2c placeholder: margin evidence for flips.}}
    \caption{Margin evidence: flip rate by margin bin.}
  \end{subfigure}
  \caption{\textbf{Joint scaling exposes a selective reliability failure and its boundary diagnostic.} All panels use the validation collection at the 10M token budget. \textbf{(a)} Along the practical width/$k$ scaling path, sensitivity falls for rare features while common features remain comparatively stable; the medium bucket is shown as a reference. \textbf{(b)} The same path crowds the selection boundary, with the share of near-cutoff examples rising from 3.7\% to 10.3\%, 26.7\%, and 54.0\% across the four displayed settings. \textbf{(c)} Active margin bins predict actual source--paraphrase flips of the active set; instances closest to the cutoff flip most often.}
  \label{fig:width-margin}
\end{figure*}

The 10M token GPT-2 results support this diagnosis. At width 12288 with $k=128$, 54.0\% of evaluation examples have $m<0.01$, compared with 10.3\% at width 3072. Moreover, active feature instances closest to the cutoff have the highest paraphrase flip rates, as shown in Figure~\ref{fig:width-margin}b--c. Thin margins therefore do not merely accompany lower aggregate sensitivity; they identify the specific features most likely to disappear from the paraphrase top-$k$ set.

The diagnosis also has a threshold-free form. Ranking (instance, feature) pairs by margin and scoring the ranking, the active margin predicts paraphrase dropout at AUROC $0.728$ on GPT-2 layer 8, and the drop rate falls monotonically across margin bins, from $94.1\%$ nearest the cutoff to $0.6\%$ farthest from it (Table~\ref{tab:fullscale-margin-flip}). The relation is not specific to one layer: across six additional GPT-2 depths the AUROC stays between 0.747 and 0.850 and declines smoothly with depth (Table~\ref{tab:margin-auroc-depth}). Because the statistic ranks pairs within a model, it needs no cross-model score scale.

We further test the margin account at the level of individual active features. For each source--paraphrase pair we record each active feature's source margin above the cutoff, its source rank, its paraphrase rank, and whether it drops. Grouping instances by distance to the cutoff gives a monotone boundary-crossing pattern: the closest instances flip most often and move toward or below the cutoff, while distant instances rarely drop. The margin therefore connects the aggregate sensitivity gap to the specific flips that produce it.

\begin{table}[!t]
\centering
\caption{\textbf{Depth check for the threshold-free margin statistic.} Each column trains a width-12288, $k=128$ TopK SAE at that GPT-2 layer and scores active margin against the paraphrase dropout label by AUROC. All values use the validation collection and the main paraphrase protocol; the layer-8 entry is the main-text anchor and is a single-seed estimate, the others are two-seed means.}
\label{tab:margin-auroc-depth}
\begingroup
\setlength{\tabcolsep}{4pt}
\renewcommand{\arraystretch}{1.05}
\fittable{%
\begin{tabular}{@{}lccccccc@{}}
\toprule
\bfseries
Layer & \bfseries 4 & \bfseries 6 & \bfseries 7 & \bfseries 8 & \bfseries 9 & \bfseries 10 & \bfseries 11 \\
\midrule
\normalfont
AUROC $\uparrow$ & 0.850 & 0.837 & 0.802 & 0.728 & 0.774 & 0.760 & 0.747 \\
\bottomrule
\end{tabular}}
\endgroup
\end{table}

Together these results trace rare-feature instability in wide TopK SAEs to the selection boundary: a paraphrase need not erase the relevant semantic representation to change the active set, it only has to perturb a near-boundary feature below the threshold for selection. That gives the repair a concrete target: keep source-active features ranked above their paraphrase-side competitors at the cutoff.

\FloatBarrier

\section{Rank Stabilization from the Margin Diagnosis}\label{sec:intervention-cost-scope}

The diagnosis identifies a concrete failure mode: rare features active on source inputs are often too close to hard paraphrase competitors near the TopK cutoff. This section asks whether that diagnosis can guide training. We compare generic consistency regularization, boundary thickening, boundary-local controls, and a targeted pairwise rank stabilization loss. The central result is that rank stabilization gives the largest rare-feature reliability gain on the validation collection while keeping reconstruction MSE and alive-feature coverage near the baseline.

\subsection{From overall agreement to boundary ordering}

A natural response is to train SAEs to produce similar feature activations on paraphrase pairs. We use \textbf{paraphrase consistency} for this family of baselines: it encourages source and paraphrase active-set masks to agree, measured over the whole mask. Table~\ref{tab:main-core-result} reports the full-scale control under the 10M protocol. Overall agreement and boundary ordering separate cleanly: a thin gap between the $k$th and $(k+1)$th features can be flipped whatever the mask overlap, which is why the repair is defined on the ordering at the cutoff. Mask agreement is also the wrong target: two active sets can overlap heavily and still disagree about the feature a downstream audit follows.

\subsection{Pairwise rank stabilization targets the diagnosed failure}

Because the failure is a discrete boundary crossing problem, a successful repair keeps source-active features above their paraphrase competitors at the boundary. Algorithm~\ref{alg:rank-stability-sae} implements this idea. In the paired branch, each source or paraphrase text is encoded by GPT-2 at layer 8 with maximum length 64, and the retained token states are mean pooled into one 768 dimensional activation vector; the reconstruction branch still trains on token-level activations. For features active on a source input, the loss penalizes hard negatives from the paraphrase that sit just below the TopK boundary and would otherwise outrank the same feature after rephrasing. In Algorithm~\ref{alg:rank-stability-sae}, the band size $R$ defines a narrow competitive band just below the cutoff, and $B_R(x')$ is a centered log-sum-exp aggregation of the hard-negative scores in that band, computed at temperature $\tau$ (with $\tau=1$ in the main setting). The standard reconstruction objective continues to train the SAE dictionary alongside the ordering term. We evaluate the gain by frequency bucket, with the largest effect expected for rare features because that is where the boundary failure concentrates.

\begin{algorithm}[H]
\caption{Pairwise rank stabilization}
\label{alg:rank-stability-sae}
\begin{algorithmic}[1]
\Require Token activations $z$, source--paraphrase activation pairs $(x,x')$, TopK scores $a(\cdot)$, band size $R$, temperature $\tau$, target margin $\gamma$, weight $\lambda$
\For{each training minibatch}
  \State Compute the standard SAE reconstruction loss $\mathcal{L}_{\mathrm{rec}}$ on token activations $z$
  \State Sample pair $(x,x')$; let $S(x)=\{i\in\operatorname{TopK}(a(x)):z_i(x)>0\}$ be the source active feature set
  \State Let $\mathcal{H}_R(x')$ be the features ranked $k+1$ through $k+R$ by paraphrase scores $a(x')$
  \State $B_R(x')\gets \tau\!\left(\log\sum_{j\in\mathcal{H}_R(x')}\exp(a_j(x')/\tau)-\log R\right)$
  \State $\mathcal{L}_{\mathrm{rank}}\gets \lambda\,\mathbb{E}_{i\in S(x)}\!\left[\gamma-\left(a_i(x')-B_R(x')\right)\right]_+$
  \State Update SAE parameters using $\mathcal{L}_{\mathrm{rec}}+\mathcal{L}_{\mathrm{rank}}$
\EndFor
\end{algorithmic}
\end{algorithm}

\subsection{Main repair result}

Table~\ref{tab:main-core-result} gives the main result on the validation collection, at width 12288, $k=128$, under the 10M token budget. The objectives differ in the pressure they put on the reconstruction loss: paraphrase consistency and contrastive alignment ask paired views to agree, while global margin, margin + paraphrase consistency, hinge boundary, LSE boundary, and rare-band margin press on the top-$k$ gap, globally, through a hard-negative band, or on rare features only. Baseline rare-feature sensitivity is $0.5846$, with $54.0\%$ of evaluation examples near the cutoff. Pulling paired activations together is not enough: contrastive alignment reaches $0.5048$ and paraphrase consistency, which raises overall active-set agreement, reaches $0.6552$. Pressing on the boundary directly is costly: global margin lowers the near-cutoff share to $26.8\%$, but reconstruction MSE rises from $0.880$ to $3.584$ and alive coverage falls from $96.70\%$ to $73.21\%$, and the band- and rare-only variants leave sensitivity within noise. Rank stabilization gives the largest rare-feature gain, from $0.5846$ to $0.7073$ ($+12.3$pp), with reconstruction MSE ($0.880\!\rightarrow\!0.888$) and alive coverage ($96.70\%\!\rightarrow\!96.66\%$) close to baseline. The gain comes from the ordering at the boundary rather than from clearing it: the near-cutoff share is essentially unchanged ($54.4\%$ vs.\ $54.0\%$), and what changes is which features are retained; Table~\ref{tab:w1-qualitative-audit} gives representative cases: the audit gathers 49 rare drop cases against 20 retained comparisons, at a median source margin of $0.285$ and a median paraphrase rank of 158. In one case a rare feature selected at rank 128, with an active margin of only $0.022$ above the cutoff, falls to rank 155 after paraphrasing and leaves the active set, while the paraphrase preserves the sentence's content.

\begin{table*}[t]
\centering
\caption{\textbf{Repair comparison and independent confirmation.} Entries are mean$\pm$SEM over five independently trained seeds; $\Delta$ Rare is in percentage points against the baseline of the same collection, and near-cutoff is the share of examples with $m<0.01$.}
\label{tab:main-core-result}
\begingroup
\setlength{\tabcolsep}{4pt}
\renewcommand{\arraystretch}{1.15}
\fittable{%
\begin{tabular}{@{}llccccc@{}}
\toprule
\bfseries
Collection & \bfseries Method & \bfseries Rare $\uparrow$ & \bfseries $\Delta$ Rare & \bfseries Near-cutoff $\downarrow$ & \bfseries MSE $\downarrow$ & \bfseries Alive $\uparrow$ \\
\midrule
\normalfont
\multirow{9}{*}{\shortstack[l]{\emph{Validation}\\(148 pairs)}} & baseline & 0.585$\pm$0.017 & +0.0 & 54.0$\pm$0.2 & 0.880$\pm$0.002 & 96.70$\pm$0.17 \\
 & paraphrase consistency & 0.655$\pm$0.008 & +7.1 & 53.7$\pm$0.2 & 0.886$\pm$0.008 & 96.58$\pm$0.19 \\
 & contrastive alignment & 0.505$\pm$0.009 & -8.0 & 54.0$\pm$0.4 & 0.881$\pm$0.004 & 96.78$\pm$0.18 \\
 & global margin & 0.428$\pm$0.049 & -15.6 & 26.8$\pm$0.3 & 3.584$\pm$0.020 & 73.21$\pm$0.22 \\
 & margin + paraphrase consistency & 0.432$\pm$0.025 & -15.2 & 27.4$\pm$0.2 & 3.566$\pm$0.036 & 73.38$\pm$0.23 \\
 & hinge boundary & 0.580$\pm$0.020 & -0.5 & 54.2$\pm$0.2 & 0.889$\pm$0.012 & 96.60$\pm$0.12 \\
 & LSE boundary & 0.586$\pm$0.016 & +0.2 & 54.0$\pm$0.3 & 0.888$\pm$0.010 & 96.65$\pm$0.14 \\
 & rare-band margin & 0.562$\pm$0.015 & -2.3 & 52.2$\pm$0.2 & 0.819$\pm$0.007 & 97.28$\pm$0.14 \\
 & \textbf{pairwise rank stabilization (ours)} & \textbf{0.707$\pm$0.013} & \textbf{+12.3} & 54.4$\pm$0.3 & 0.888$\pm$0.011 & 96.66$\pm$0.14 \\
\midrule
\multirow{2}{*}{\shortstack[l]{\emph{Held-out}\\(1712 pairs)}} & baseline & 0.467$\pm$0.007 & +0.0 & 54.0$\pm$0.2 & 0.880$\pm$0.002 & 96.70$\pm$0.17 \\
 & \textbf{pairwise rank stabilization (ours)} & \textbf{0.555$\pm$0.011} & \textbf{+8.8} & 54.4$\pm$0.3 & 0.888$\pm$0.011 & 96.66$\pm$0.14 \\
\bottomrule
\end{tabular}}
\endgroup
\end{table*}

Because the comparison and the choice of $\lambda$ and $R$ are made on the validation collection, we freeze the selected configuration ($\lambda=0.004$, $R=8$) and evaluate it forward-only on the held-out collection: 856 source texts (1712 pairs). Across five seeds, rare-feature sensitivity rises from $0.467\pm0.007$ to $0.555\pm0.011$, a gain of $+8.83$pp, with mean reconstruction MSE moving from $0.880$ to $0.888$ and mean alive coverage from $96.70\%$ to $96.66\%$ (Table~\ref{tab:main-core-result}, lower block). Both blocks score the same trained checkpoints, so reconstruction, alive coverage, and the near-cutoff share are model properties that agree across them; only the sensitivity columns respond to the collection. Over the validation grid the same objective improves rare-feature sensitivity in all sixteen configurations, from $+4.9$ to $+12.3$pp (Table~\ref{tab:fullscale-rank-ablation}), so the held-out estimate sits inside the range the grid already spans.

\begin{table}[!t]
\centering
\caption{\textbf{Rank stability sweep over $\lambda$ and $R$.} Rare-feature sensitivity (mean over five seeds) at width 12288, $k=128$, and the 10M token budget, on the validation collection; the value after $\uparrow$ gives the change in percentage points against its $0.5846$ baseline.}
\label{tab:fullscale-rank-ablation}
\begingroup
\renewcommand{\arraystretch}{1.05}
\fittable{%
\def\tabfontsize{8pt}    
\def\tabbaselineskip{10pt}
\def\tabcolgap{7pt}      
\fontsize{\tabfontsize}{\tabbaselineskip}\selectfont
\setlength{\tabcolsep}{\tabcolgap}
\begin{tabular}{@{}llllr@{}}
\toprule
\bfseries
$\boldsymbol{\lambda \backslash R}$ & \bfseries $\mathbf{R}=8$ & \bfseries $\mathbf{R}=16$ & \bfseries $\mathbf{R}=32$ & \bfseries $\mathbf{R}=64$ \\
\midrule
\normalfont
$5\times10^{-4}$ & 0.640 $\uparrow$5.5 & 0.637 $\uparrow$5.2 & 0.634 $\uparrow$4.9 & 0.635 $\uparrow$5.0 \\
$1\times10^{-3}$        & 0.689 $\uparrow$10.4 & 0.663 $\uparrow$7.8 & 0.656 $\uparrow$7.1 & 0.675 $\uparrow$9.1 \\
$2\times10^{-3}$ & 0.680 $\uparrow$9.6 & 0.697 $\uparrow$11.2 & 0.689 $\uparrow$10.4 & 0.684 $\uparrow$9.9 \\
$4\times10^{-3}$ & \textbf{0.707 $\uparrow$12.3} & 0.679 $\uparrow$9.5 & 0.697 $\uparrow$11.3 & 0.687 $\uparrow$10.3 \\
\bottomrule
\end{tabular}}
\endgroup
\end{table}

\subsection{Robustness, cost, and scope}
\label{sec:robustness-scope}

Moving the boundary carries a cost: the margin comparison lowers the share of near-cutoff examples and pays for it in reconstruction and alive-feature coverage (Appendix Table~\ref{tab:fullscale-margin-repair}). The gain is not tied to the token budget: at 200k, 1M, and 10M tokens the wide baseline remains less sensitive to rare features than the narrower one, and margin pressure is lower still (Appendix Figure~\ref{fig:cost-severity}a).

The same setting transfers beyond GPT-2. Table~\ref{tab:crossmodel-fullscale} reads each model within itself under the same split and evaluation procedure. Qwen has a crowded boundary (74.4\% near-cutoff examples) and rank stabilization raises rare-feature sensitivity from $0.929$ to $0.946$ without moving reconstruction or alive coverage; Gemma's boundary is sparser (18.6\%) with much lower alive coverage, and the same objective is near neutral there ($-0.8$pp). The near-cutoff share therefore reports how much room a boundary repair has to act in a given model.
Two SAE-internal summaries of the active set move in the same direction: retention on the true paraphrase rises from $0.690$ to $0.755$ and the paraphrase/source effect ratio from $0.735$ to $0.791$, while affinity for a mismatched paraphrase rises by less, leaving the retention gap essentially unchanged (0.238 to 0.248; Appendix Table~\ref{tab:feature-usability-checks}).

\begin{table}[t]
\centering
\caption{\textbf{Model family checks.} GPT-2 (the main setting), Qwen2.5-1.5B-Instruct (written Qwen2.5-1.5B below), and Gemma-2-9B use per-token active budgets and source-disjoint paraphrase evaluation at the 10M token budget; every row is the mean over three independently trained seeds, except the GPT-2 reference rows, which are the five-seed values of Table~\ref{tab:main-core-result}. \emph{$W/k$} is dictionary width over the active budget; \emph{Rare} is rare-feature sensitivity and \emph{$\Delta$ Rare} its change in percentage points against the baseline of the same model and width (\emph{base} rows are those baselines); \emph{Near-cutoff} and \emph{Alive} are percentages and \emph{MSE} is reconstruction error.}
\label{tab:crossmodel-fullscale}
\begingroup
\setlength{\tabcolsep}{3.9pt}
\renewcommand{\arraystretch}{1.15}
\fittable{%
\begin{tabular}{@{}llcrrrrr@{}}
\toprule
\bfseries
Model & \bfseries $W/k$ & \bfseries Run & \bfseries Rare $\uparrow$ & \bfseries $\Delta$ Rare & \bfseries Near-cutoff $\downarrow$ & \bfseries MSE $\downarrow$ & \bfseries Alive $\uparrow$ \\
\midrule
\normalfont
GPT-2 & 3072/32 & base & 0.650$\pm$0.015 & --- & 10.3$\pm$0.2 & 1.672$\pm$0.002 & 98.8$\pm$0.1 \\
GPT-2 & 12288/128 & base & 0.585$\pm$0.017 & --- & 54.0$\pm$0.2 & 0.880$\pm$0.002 & 96.7$\pm$0.2 \\
GPT-2 & 12288/128 & rank & 0.707$\pm$0.013 & $+12.3$ & 54.4$\pm$0.3 & 0.888$\pm$0.011 & 96.7$\pm$0.1 \\
\midrule
Gemma-2-9B & 16384/64 & base & 0.672$\pm$0.010 & --- & 18.5$\pm$0.4 & 1.440$\pm$0.002 & 76.4$\pm$0.5 \\
Gemma-2-9B & 65536/64 & base & 0.647$\pm$0.004 & --- & 18.6$\pm$0.1 & 1.339$\pm$0.000 & 51.5$\pm$0.2 \\
Gemma-2-9B & 65536/64 & rank & 0.639$\pm$0.007 & $-0.8$ & 18.1$\pm$0.2 & 1.339$\pm$0.001 & 51.3$\pm$0.2 \\
Qwen2.5-1.5B & 6144/32 & base & 0.987$\pm$0.010 & --- & 17.8$\pm$0.1 & 0.596$\pm$0.013 & 78.8$\pm$0.3 \\
Qwen2.5-1.5B & 24576/128 & base & 0.929$\pm$0.003 & --- & 74.4$\pm$0.6 & 0.397$\pm$0.020 & 86.8$\pm$0.1 \\
Qwen2.5-1.5B & 24576/128 & rank & 0.946$\pm$0.008 & $+1.7$ & 74.1$\pm$0.3 & 0.382$\pm$0.013 & 85.8$\pm$0.3 \\
\bottomrule
\end{tabular}}
\endgroup
\end{table}

\section{Discussion and Limitations}

\paragraph{Scope and interpretation.}
Wider dictionaries expose useful low frequency structure, and the reliability cost we study appears in the TopK regime, where competition for the active set gives rare features thinner selection margins and makes them easier to dislodge under meaning preserving rewrites. The rare/medium/common buckets are a diagnostic partition by frequency, not a semantic taxonomy. Paraphrase sensitivity probes whether an interpreted feature remains the same unit under ordinary changes in surface form, one prerequisite for the explanation, circuit-analysis, and intervention uses SAEs are built for. Our evidence concerns activation reliability, leaving two open questions: whether dropped features represent lost human-interpretable concepts, and whether these gains improve downstream interpretability. Recent evaluations raise the same concerns~\citep{chanin2026synthsaebench,korznikov2026sanitychecks,arora2026neuronbasis,patel2026uncertaintycorrectness}.

\paragraph{Repair and reporting.}
Global margin pressure is a mechanistic reference point: it moves the boundary at a reconstruction and coverage cost, showing the boundary is movable and pricing the move. Rank stabilization is the targeted alternative, and the practical recommendation is to report feature sensitivity alongside reconstruction and sparsity. Reporting it costs one paraphrase pass per evaluation set, and it is the only measurement we ran that separates two checkpoints with near-identical reconstruction, sparsity, and alive-feature counts.

\section{Conclusion}

SAE scaling aims to recover finer latent structure from language model activations. Our results show that this promise should be paired with a reliability check: a feature that appears interpretable on top activating examples may still fail to remain available under meaning preserving rewrites. In TopK SAEs, this failure is systematic: it concentrates in rare features, follows the geometry of the selection boundary, and can be predicted by margins near the top-$k$ cutoff. Pairwise rank stabilization shows how such a diagnosis guides repair: it targets the source--paraphrase ordering failure behind the drops. SAE evaluation should therefore report feature reliability alongside reconstruction and feature count, since the two can move in opposite directions over the same scaling path.

\FloatBarrier
\bibliographystyle{iclr2027_conference}
\bibliography{references}

\clearpage
\appendix
\setcounter{table}{0}
\renewcommand{\thetable}{A.\arabic{table}}
\setcounter{algorithm}{0}
\renewcommand{\thealgorithm}{A.\arabic{algorithm}}

\section{Additional Experimental Details}

\paragraph{Appendix organization.}
The appendix gives the experimental protocol behind the main claims: the main setting, the paraphrase procedure, the qualitative drop audit, the statistical unit and compute setup, a boundary derivation of the margin diagnostic, the per-seed diagnosis and repair tables, the token-budget and model-family scope checks, and the specificity, loss-variant, and usability controls.

\paragraph{Main setting.}
Table~\ref{tab:setup-overview} records the main GPT-2 configuration behind the primary claims.

\begin{table}[H]
\centering
\caption{\textbf{Main GPT-2 experimental setup.} The primary claims use token budgets and evaluate paraphrase sensitivity on source disjoint text pairs, so the reliability probe is separated from SAE training, frequency estimation, source selection, and auxiliary pair training. Model comparison and hyperparameter selection use the validation collection; the independent confirmation uses the held-out collection.}
\label{tab:setup-overview}
\begingroup
\setlength{\tabcolsep}{6pt}
\renewcommand{\arraystretch}{1.2}
\fittable{%
\begin{tabular}{@{}p{0.18\linewidth}>{\columncolor{gray!15}}p{0.77\linewidth}}
\toprule
\bfseries
Component & \bfseries Setting \\
\arrayrulecolor{black}\hdashline
\normalfont
Model / hook & GPT-2 small~\citep{radford2019language}, layer 8 residual stream; hook \texttt{blocks.8.hook\_resid\_pre} \\
\arrayrulecolor{white}\hdashline
Activation corpus & \texttt{Skylion007/openwebtext}~\citep{gokaslan2019openwebtext}; training split; 64-token contexts \\
\arrayrulecolor{white}\hdashline
Token budget & 10M token activations primary; 200k and 1M robustness checks \\
\arrayrulecolor{white}\hdashline
Data partitions & deterministic train, frequency, source selection, and evaluation partitions; paraphrase pairs are disjoint by original source text across the validation and held-out collections \\
\arrayrulecolor{white}\hdashline
Main widths & 3072 / $k=32$ and 12288 / $k=128$ TopK SAEs; summaries over independent seeds \\
\arrayrulecolor{white}\hdashline
Evaluation & feature bucket sensitivity on source text disjoint paraphrase pairs; model comparison on the validation collection, independent confirmation on the held-out collection \\
\arrayrulecolor{black}\bottomrule
\end{tabular}}
\endgroup
\end{table}

\paragraph{Paraphrase procedure.}
We use paraphrases to test meaning-preserving feature availability while keeping held-out source texts separate from auxiliary pair training. Paraphrase pairs are generated with GPT-4o at temperature 0.7, with a 256-token output cap, using the instruction: ``Rewrite the following text as a paraphrase. Keep the same meaning but use different words and sentence structure. Keep a similar length. Output only the paraphrase.'' Generated pairs are manually reviewed for meaning preservation before being used in the reliability evaluation. The main GPT-2 pair collection starts from 150 source texts with two paraphrases per text and contains 295 retained source--paraphrase pairs after generation and filtering. When the same collection is used for both auxiliary pair training and evaluation, we split by original source text with a 0.5 evaluation fraction and split seed 20260504, yielding approximately 147 auxiliary pairs and 148 held-out evaluation pairs; the latter form the validation collection of the main text. Source and paraphrase texts are run through GPT-2 with maximum sequence length 64; we collect the layer 8 residual stream and mean pool over the retained token positions to obtain one 768-dimensional activation vector for each side of the pair. Main sensitivity evaluations sample activating texts for selected features and generate three paraphrases per text. We report sensitivity by feature bucket; individual paraphrases contribute to the within-source estimate rather than being treated as independent evaluation samples.

\paragraph{Qualitative paraphrase drop audit.}
To avoid overinterpreting sensitivity as a semantic ground truth label, we build a qualitative audit set from the margin diagnostic. Each example attaches source/paraphrase text to a rare feature that is selected in the source top-$k$ set and leaves the paraphrase top-$k$ set, along with source rank, paraphrase rank, and active margin. We score activation by the post-ReLU sign $f>0$ rather than by top-$k$ membership, because the two differ near the boundary: a feature can enter the top-$k$ set while its encoder score is non-positive, in which case the ReLU zeroes it and it was never active. A substantial share of top-$k$ drop candidates have such non-positive source scores, so the table below illustrates the selection-boundary event of Section~\ref{sec:preliminaries} and is not a quantitative failure rate. The audit sample contains 49 deduplicated rare drop cases and 20 rare retained comparisons; the rare drop cases have median source active margin 0.285 and median paraphrase rank 158. Table~\ref{tab:w1-qualitative-audit} gives representative examples.

\begin{table}[H]
\centering
\caption{\textbf{Qualitative examples of paraphrase drops for rare features.} Each row is a rare feature that is selected in the source top-$k$ set and is not selected in the paraphrase top-$k$ set. This is the top-$k$ selection criterion, which is weaker than the $z>0$ activation criterion used for the sensitivity statistic: the source-side encoder score is often non-positive, so a row records a selection-boundary event rather than a confirmed loss of an active feature. \emph{Feature} is the dictionary index, \emph{Sour.\ rank} and \emph{Para.\ rank} are its ranks in the source and paraphrase top-$k$ sets, and \emph{Margin} is the source active margin. Examples are drawn from the audit sample described in the text. The paraphrase preserves recognizable sentence level content while the feature no longer survives the boundary.}
\label{tab:w1-qualitative-audit}
\begingroup
\setlength{\tabcolsep}{2pt}
\renewcommand{\arraystretch}{1.12}
\fittable{%
\begin{tabular}{>{\columncolor{gray!12}}p{0.50\linewidth}>{\centering\arraybackslash}m{0.09\linewidth}>{\centering\arraybackslash}m{0.12\linewidth}>{\centering\arraybackslash}m{0.13\linewidth}>{\centering\arraybackslash}m{0.08\linewidth}}
\toprule
\textbf{Example} & \textbf{Feature} & \textbf{Sour.\ rank} & \textbf{Para.\ rank} & \textbf{Margin} \\
\arrayrulecolor{black}\hdashline
\textbf{Source:} Former secretary of state Hillary Clinton meets voters at a campaign rally in St.\par
\textbf{Paraphrase:} Hillary Clinton, the ex-secretary of state, engages with voters during a rally in St. & 10520 & 128 & 200 & 0.017 \\
\arrayrulecolor{white}\hdashline
\textbf{Source:} Ad blockers are often painted as the enemy of online publishers, but sometimes things are more complicated. AdBlock Plus, for example, just announced...\par
\textbf{Paraphrase:} While ad blockers are frequently seen as adversaries by online publishers, the situation can be more nuanced. For instance, AdBlock Plus recently rev... & 9241 & 128 & 134 & 0.018 \\
\arrayrulecolor{white}\hdashline
\textbf{Source:} BIGBANG is one of those musical entities that transcends language.\par
\textbf{Paraphrase:} BIGBANG is a musical group that surpasses language barriers. & 335 & 128 & 155 & 0.022 \\
\arrayrulecolor{white}\hdashline
\textbf{Source:} The opinions expressed by columnists are their own and do not represent the views of Townhall.com. You have to give President Barack Obama credit for...\par
\textbf{Paraphrase:} The views put forth by columnists are solely theirs and do not reflect those of Townhall.com. One must acknowledge President Barack Obama for his con... & 809 & 128 & 223 & 0.030 \\
\arrayrulecolor{white}\hdashline
\textbf{Source:} Story highlights Tyka Nelson says her brother's favorite color was ...\par
\textbf{Paraphrase:} Tyka Nelson reveals that her brother's most cherished color was ... & 1397 & 128 & 205 & 0.035 \\
\arrayrulecolor{black}\bottomrule
\end{tabular}}
\endgroup
\end{table}

\paragraph{Statistical unit.}
All means and tests in the appendix follow the same convention as the main text: independent SAE training seeds are the statistical unit, error terms are SEM over seeds, and pairwise tests are two sided Welch tests unless otherwise noted.

\paragraph{SAE optimization details.}
The main GPT-2 width 12288, $k=128$ runs use Adam with learning rate $3\times 10^{-4}$, batch size 256, and at most 20k optimization steps. The rank stability runs use the same reconstruction objective and add the pairwise ordering term with target margin $\gamma=0.01$; Table~\ref{tab:main-core-result} reports the configuration selected on the validation collection, $\lambda=0.004$ and $R=8$, and Appendix Table~\ref{tab:fullscale-rank-ablation} reports the full $\lambda\times R$ sweep. Margin pressure comparisons use $\lambda=0.01$ and target margin $0.01$. Each run records its exact command, split, and code revision.

\paragraph{Compute resources.}
Every reported run is a single-GPU job and fits on one NVIDIA A100-80GB. For the largest Qwen and Gemma checks, activation collection first loads the corresponding base language model on a single high memory node; the subsequent SAE fits, reliability evaluations, summary tables, and figures operate on stored activations. The anonymized release package records the run commands, code hashes, and output paths needed to reproduce the reported tables and figures.

\paragraph{Existing assets and licenses.}
The experiments use publicly released pretrained language models and associated tokenizers/weights from their original release channels or HuggingFace model cards, together with standard open source Python libraries. We do not redistribute a new pretrained language model or a scraped text dataset. The anonymized release will list exact model identifiers, upstream citations, license or terms of use strings, and package versions for GPT-2, Qwen, Gemma, and the software dependencies used to reproduce the paper.

\subsection{Full GPT-2 Evidence}
\label{app:full-gpt2}

This subsection carries the GPT-2 evidence behind Figure~\ref{fig:width-margin} and Table~\ref{tab:main-core-result}. Table~\ref{tab:denominators} fixes what the sensitivity statistic averages over, so that each rate below can be read against the number of feature instances behind it: on the validation collection a source text carries 131 rare active features on average and the rare-feature rate averages over 918 source-side instances, while the frequency partition that defines the buckets is estimated separately on a 1M-activation sample.

Three of the remaining tables test the diagnosis away from the single main setting: Table~\ref{tab:fullscale-width-sweep} repeats the width sweep at 200k, 1M, and 10M tokens, and Table~\ref{tab:fullscale-margin-flip} shows that the active features nearest the source cutoff are the ones paraphrases displace. The other two support the repair: Table~\ref{tab:fullscale-margin-repair} prices global boundary pressure as the mechanistic reference point, and Table~\ref{tab:fullscale-rank-ablation} reports the $\lambda\times R$ sweep that selects the main configuration.

\begin{table}[!htbp]
\centering
\caption{\textbf{Denominators behind the rare-feature sensitivity statistic.} GPT-2 width 12288, $k=128$, 10M token budget, validation collection. Feature counts are estimated on the separate 1M-activation frequency partition; the evaluation counts are the pairs the sensitivity statistic is averaged over. \emph{Mean} is the value over the five main seeds and \emph{Range over seeds} its spread, left blank where the quantity does not vary across seeds.}
\label{tab:denominators}
\begingroup
\setlength{\tabcolsep}{6pt}
\renewcommand{\arraystretch}{1.06}
\fittable{%
\begin{tabular}{@{}lrr@{}}
\toprule
\bfseries
Quantity & \bfseries Mean & \bfseries Range over seeds \\
\midrule
\normalfont
Rare features active on a source text & 131 & 120--141 \\
Rare active feature instances (source side) & 918 & 760--1052 \\
\midrule
Paraphrase evaluation pairs & 148 & --- \\
Paraphrase pairs, all rows & 295 & --- \\
Frequency partition activation count & $1{,}000{,}000$ & --- \\
Rare features, frequency partition & 2599 & --- \\
Medium features, frequency partition & 511 & --- \\
Common features, frequency partition & 24 & --- \\
Alive features, of 12288 & 11892 & --- \\
\bottomrule
\end{tabular}}
\endgroup
\end{table}

\begin{table}[t]
\centering
\caption{\textbf{GPT-2 width sweep on the validation collection, under token budgets.} Values are mean$\pm$SEM over independently trained SAEs, with $n$ the number of seeds; \emph{Rare}, \emph{Medium}, and \emph{Common} are sensitivity for each feature-frequency bucket, \emph{Near-cutoff} is the percentage of evaluation examples with $m<0.01$, \emph{MSE} is reconstruction error, and \emph{Alive} is alive-feature coverage in percent. The sensitivity drop is concentrated in rare features; common features remain stable across the same sweep.}
\label{tab:fullscale-width-sweep}
\begingroup
\setlength{\tabcolsep}{2.9pt}
\renewcommand{\arraystretch}{1.18}
\fittable{%
\begin{tabular}{@{}lrrrrrrrrr@{}}
\toprule
\bfseries
Tokens & \bfseries Width & \bfseries $k$ & \bfseries $n$ & \bfseries Rare $\uparrow$ & \bfseries Medium $\uparrow$ & \bfseries Common $\uparrow$ & \bfseries Near-cutoff $\downarrow$ & \bfseries MSE $\downarrow$ & \bfseries Alive $\uparrow$ \\
\midrule
\normalfont
10M & 768 & 16 & 1 & 0.795$\pm$0.000 & 0.519$\pm$0.000 & 0.993$\pm$0.000 & 3.7$\pm$0.0 & 2.786$\pm$0.000 & 97.1$\pm$0.0 \\
10M & 3072 & 32 & 5 & 0.650$\pm$0.015 & 0.683$\pm$0.031 & 0.998$\pm$0.002 & 10.3$\pm$0.2 & 1.672$\pm$0.002 & 98.8$\pm$0.1 \\
10M & 12288 & 128 & 5 & 0.585$\pm$0.017 & 0.617$\pm$0.008 & 0.956$\pm$0.008 & 54.0$\pm$0.2 & 0.880$\pm$0.002 & 96.7$\pm$0.2 \\
1M & 3072 & 32 & 3 & 0.779$\pm$0.021 & 0.704$\pm$0.041 & 0.963$\pm$0.009 & 9.8$\pm$0.1 & 1.816$\pm$0.004 & 98.8$\pm$0.2 \\
1M & 12288 & 128 & 3 & 0.748$\pm$0.019 & 0.661$\pm$0.013 & 0.988$\pm$0.007 & 61.8$\pm$0.5 & 1.089$\pm$0.002 & 99.8$\pm$0.0 \\
200k & 3072 & 32 & 3 & 0.802$\pm$0.016 & 0.806$\pm$0.100 & 0.992$\pm$0.008 & 9.6$\pm$0.1 & 2.178$\pm$0.004 & 98.0$\pm$0.1 \\
200k & 12288 & 128 & 3 & 0.793$\pm$0.003 & 0.616$\pm$0.018 & 1.000$\pm$0.000 & 66.5$\pm$0.6 & 1.510$\pm$0.001 & 99.3$\pm$0.0 \\
\bottomrule
\end{tabular}}
\endgroup
\end{table}

\begin{table}[t]
\centering
\caption{\textbf{Global boundary pressure as a mechanistic reference point.} Margin aware training reduces the share of near-cutoff examples, showing that the boundary is manipulable, but reconstruction, alive feature coverage, and sensitivity for rare features reveal the cost of treating global clearance as the training objective. \emph{Rare}, \emph{Medium}, and \emph{Common} are sensitivity for each feature-frequency bucket; \emph{Near-cutoff} and \emph{Alive} are percentages, \emph{Margin} is the mean top-$k$ gap, \emph{MSE} is reconstruction error, and $n$ is the number of seeds. The 10M rows use the same paraphrase evaluation procedure as Table~\ref{tab:main-core-result}; smaller-budget rows are retained from the token-budget sweep.}
\label{tab:fullscale-margin-repair}
\begingroup
\setlength{\tabcolsep}{1.3pt}
\renewcommand{\arraystretch}{1.08}
\fittable{%
\begin{tabular}{@{}lrrrrrrrrr@{}}
\toprule
\bfseries
Tokens & \bfseries Method & \bfseries $n$ & \bfseries Rare $\uparrow$ & \bfseries Medium $\uparrow$ & \bfseries Common $\uparrow$ & \bfseries Near-cutoff $\downarrow$ & \bfseries Margin & \bfseries MSE $\downarrow$ & \bfseries Alive $\uparrow$ \\
\midrule
\normalfont
10M & baseline & 5 & 0.454$\pm$0.012 & 0.578$\pm$0.006 & 0.952$\pm$0.006 & 54.0$\pm$0.2 & 0.024$\pm$0.009 & 0.880$\pm$0.002 & 96.7$\pm$0.2 \\
10M & margin & 5 & 0.424$\pm$0.017 & 0.619$\pm$0.009 & 0.998$\pm$0.001 & 26.8$\pm$0.3 & 0.060$\pm$0.014 & 3.584$\pm$0.020 & 73.2$\pm$0.2 \\
1M & baseline & 3 & 0.748$\pm$0.019 & 0.661$\pm$0.013 & 0.988$\pm$0.007 & 61.8$\pm$0.5 & 0.011$\pm$0.000 & 1.089$\pm$0.002 & 99.8$\pm$0.0 \\
1M & margin & 3 & 0.495$\pm$0.027 & 0.666$\pm$0.009 & 0.982$\pm$0.003 & 36.8$\pm$0.2 & 0.041$\pm$0.009 & 3.310$\pm$0.109 & 81.8$\pm$0.1 \\
200k & baseline & 3 & 0.793$\pm$0.003 & 0.616$\pm$0.018 & 1.000$\pm$0.000 & 66.5$\pm$0.6 & 0.009$\pm$0.000 & 1.510$\pm$0.001 & 99.3$\pm$0.0 \\
200k & margin & 3 & 0.546$\pm$0.006 & 0.678$\pm$0.010 & 0.980$\pm$0.004 & 44.2$\pm$0.2 & 0.035$\pm$0.011 & 4.697$\pm$1.134 & 78.1$\pm$0.3 \\
\bottomrule
\end{tabular}}
\endgroup
\end{table}

\begin{table}[t]
\centering
\caption{\textbf{Margin evidence for paraphrase flips} on the validation collection at the 10M token budget. Active feature instances are grouped by source active margin; $n$ is the number of instances in the bin, \emph{Flip rate} is the share of those instances that leave the paraphrase active set, and \emph{Source rank} and \emph{Para.\ rank} are the mean top-$k$ rank of the feature on the source and paraphrase side. Bins near the boundary have substantially higher paraphrase drop rates, linking the aggregate sensitivity loss to concrete active set flips.}
\label{tab:fullscale-margin-flip}
\begingroup
\setlength{\tabcolsep}{3.2pt}
\renewcommand{\arraystretch}{1.08}
\fittable{%
\begin{tabular}{lrrrr}
\toprule
\bfseries
Margin bin & \bfseries $n$ & \bfseries Flip rate $\downarrow$ & \bfseries Source rank & \bfseries Para.\ rank \\
\midrule
\normalfont
$[4.89{\times}10^{-6}, 0.156]$ & 4719 & 0.941 & 120.0 & 156.3 \\
$[0.156, 0.337]$ & 4719 & 0.896 & 104.0 & 133.4 \\
$[0.337, 0.561]$ & 4721 & 0.830 & 87.9 & 119.5 \\
$[0.561, 0.875]$ & 4719 & 0.677 & 71.9 & 99.2 \\
$[0.875, 1.31]$ & 4719 & 0.447 & 56.0 & 79.7 \\
$[1.31, 2.1]$ & 4720 & 0.203 & 40.6 & 57.3 \\
$[2.1, 4.47]$ & 4719 & 0.080 & 25.6 & 37.7 \\
$[4.47, 62.3]$ & 4720 & 0.006 & 10.0 & 13.0 \\
\bottomrule
\end{tabular}}
\endgroup
\end{table}

Read together, these tables localize the reliability loss to rare features, tie it to a crowded cutoff, and identify a rank-stabilization setting that improves rare-feature sensitivity without the reconstruction and alive-coverage costs that global boundary pressure incurs. The width effect is not an artifact of a single token budget: rare-feature sensitivity at width 12288 stays below the narrower settings at 200k, 1M, and 10M tokens, while common-feature sensitivity is near ceiling in every row of Table~\ref{tab:fullscale-width-sweep}. The repair is not a lucky configuration either: all sixteen cells of the $\lambda\times R$ sweep improve rare-feature sensitivity over the $0.5846$ baseline, spanning $+4.9$ to $+12.3$ percentage points, so the selected $\lambda=0.004$, $R=8$ cell sits inside a broad basin rather than on an isolated peak. And the aggregate statistic and the instance-level event agree: Table~\ref{tab:fullscale-margin-flip} shows that active features nearest the source-side cutoff are the ones paraphrases most often displace, which is the boundary crossing the sensitivity rate counts.

\subsection{Depth Check Across GPT-2 Layers}

We also check whether the boundary signal is specific to the main GPT-2 layer-8 setting. For this depth check we train width-12288, $k=128$ TopK SAEs on layers 2, 8, and 10 using 1M-token activation caches and the same evaluation protocol, with $N=3$ independently trained baseline SAEs per layer. Table~\ref{tab:depth-sanity} shows that the magnitude varies with depth, but every inspected layer has a substantial near-cutoff share and a visible frequency-dependent sensitivity pattern; the lower rare-feature sensitivity at layer 2 sits alongside the highest near-cutoff share in the check, which is the ordering the margin account predicts. On layer 10, pairwise rank stabilization improves rare-feature sensitivity by 5.2 percentage points with negligible reconstruction and alive-feature changes. These runs use 1M-token caches rather than the 10M-token budget of the main study, which is why their absolute rare-feature sensitivity is higher than the corresponding main-table row, and the full 10M-token evidence remains centered on GPT-2 layer 8.

\begin{table}[!htbp]
\centering
\caption{\textbf{GPT-2 depth check.} Rows use 1M-token activation caches and $N=3$ independently trained baseline TopK SAEs per layer, with width 12288, $k=128$, and the same evaluation protocol throughout. \emph{Near-cutoff} is the percentage of evaluation examples with $m<0.01$; the remaining columns are rare, medium, and common feature sensitivity. Boundary crowding appears across the inspected layers, while the full 10M-token evidence remains centered on GPT-2 layer 8.}
\label{tab:depth-sanity}
\fittable{%
\begin{tabular}{@{}lrrrr@{}}
\toprule
\bfseries
Layer & \bfseries Near-cutoff $\downarrow$ & \bfseries Rare $\uparrow$ & \bfseries Medium $\uparrow$ & \bfseries Common $\uparrow$ \\
\midrule
\normalfont
2 & 69.2 & 0.606 & 0.763 & 0.976 \\
8 & 55.4 & 0.793 & 0.769 & 0.977 \\
10 & 43.8 & 0.721 & 0.798 & 1.000 \\
\bottomrule
\end{tabular}}
\end{table}
\FloatBarrier

\subsection{Token budget and model family scope}
\begin{figure*}[!htbp]
  \centering
  \begin{subfigure}[t]{0.48\textwidth}
    \centering
    \IfFileExists{figures/v4_fig4_panel_a_budget_robustness.pdf}{\includegraphics[width=\linewidth]{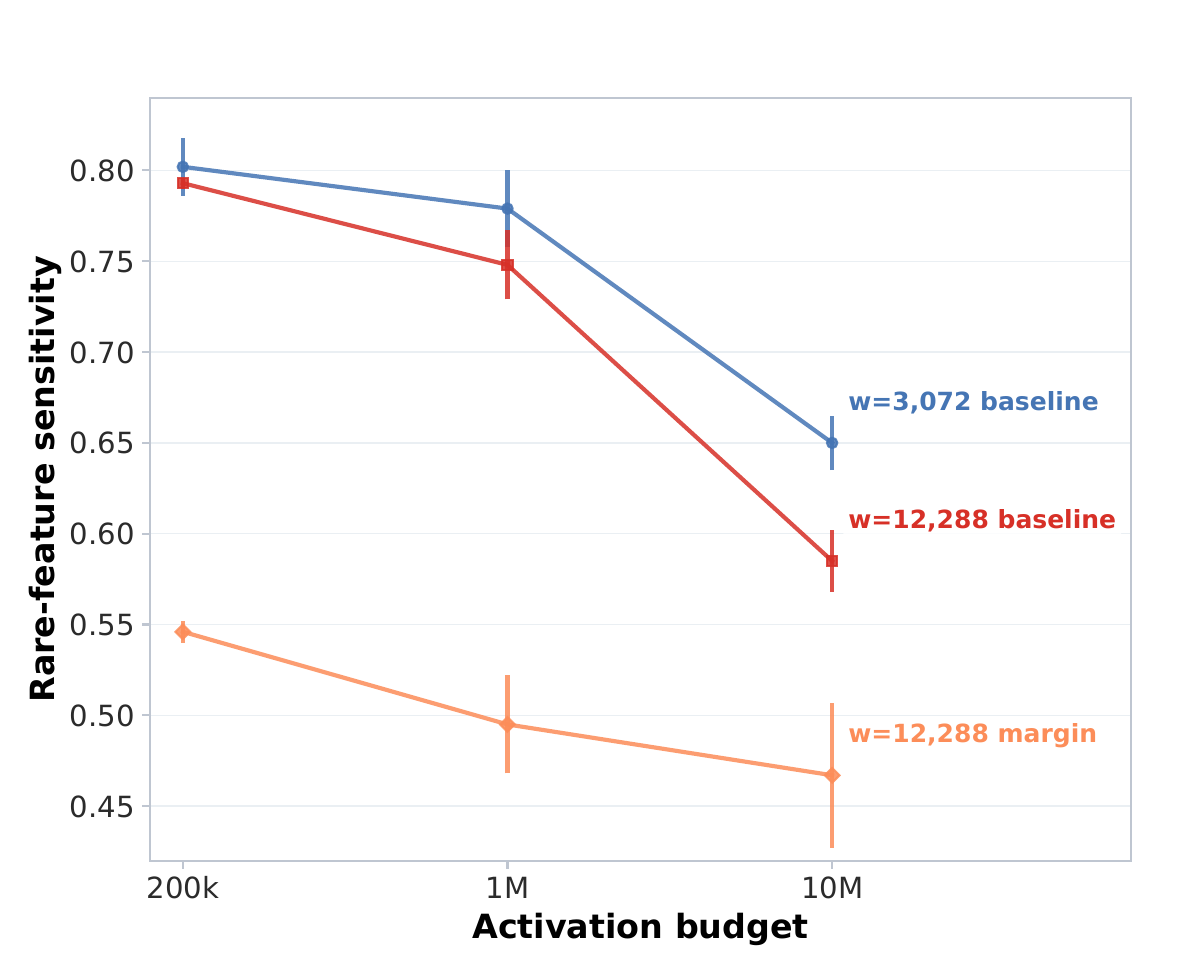}}{\figplaceholder{1.05in}{Appendix figure: token-budget robustness.}}
    \caption{Token-budget robustness: sensitivity of rare features across explicit token budgets.}
  \end{subfigure}\hfill
  \begin{subfigure}[t]{0.48\textwidth}
    \centering
    \IfFileExists{figures/v4_fig4_panel_b_stress_tradeoff.pdf}{\includegraphics[width=\linewidth]{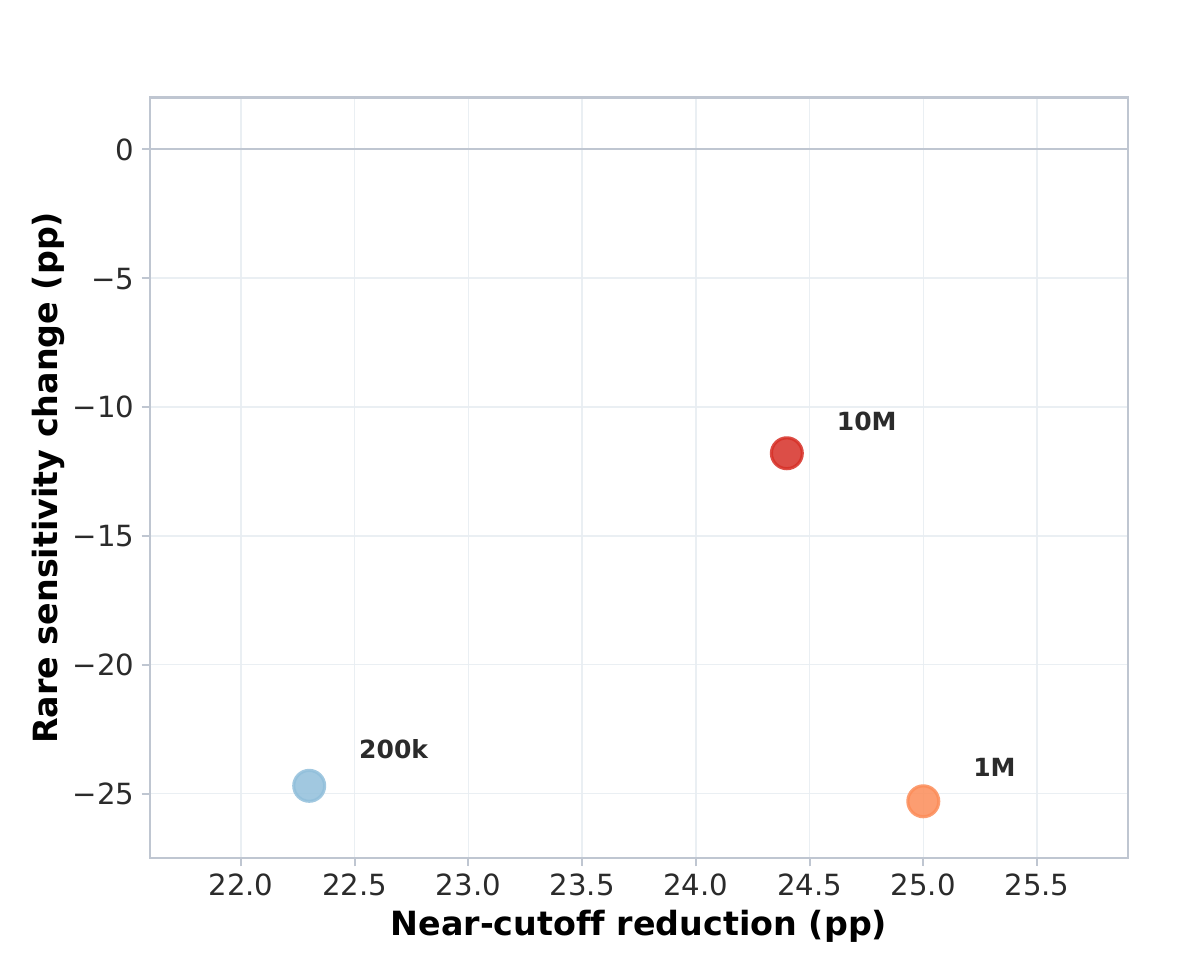}}{\figplaceholder{1.05in}{Appendix figure: boundary intervention tradeoff.}}
    \caption{Boundary intervention: movement versus change in rare sensitivity.}
  \end{subfigure}
  \vspace{0.35em}
  \begin{subfigure}[t]{0.48\textwidth}
    \centering
    \IfFileExists{figures/v4_fig4_panel_c_cross_model_severity.pdf}{\includegraphics[width=\linewidth]{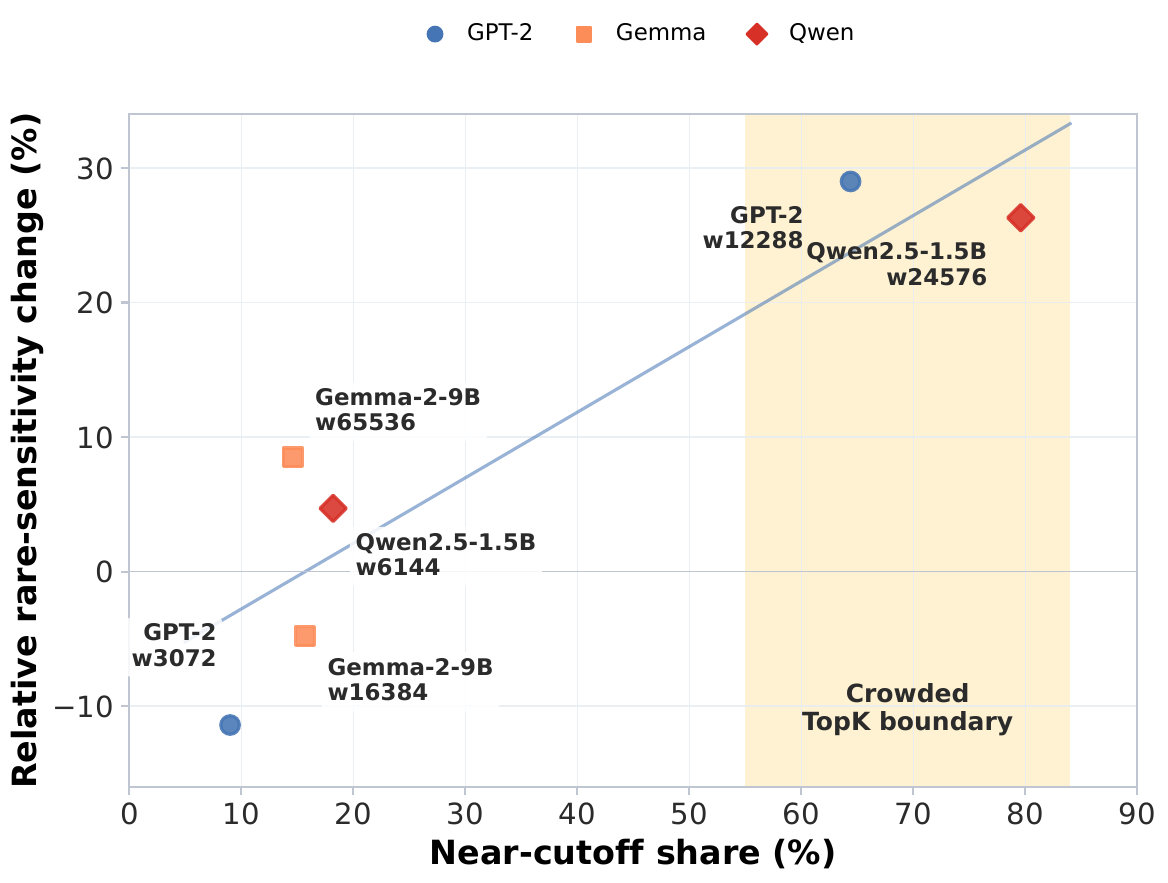}}{\figplaceholder{1.05in}{Appendix figure: model-family severity check.}}
    \caption{Qwen/Gemma check: boundary severity and repair movement.}
  \end{subfigure}
  \caption{\textbf{Token budget and model family scope checks.} \textbf{(a)} Across 200k, 1M, and 10M token budgets, the wide baseline has lower sensitivity for rare features than the narrower baseline, and the margin objective is lower still. \textbf{(b)} Each point compares the width 12288 margin objective with its same budget baseline: moving right means the share of near-cutoff examples decreases, while vertical position reports the change in rare sensitivity in percentage points. \textbf{(c)} Each model family is read within itself: the share of near-cutoff examples describes how crowded that model's own boundary is, and the repair movement is largest where that share is largest.}
  \label{fig:cost-severity}
\end{figure*}

Figure~\ref{fig:cost-severity} gives the scope checks summarized in Section~\ref{sec:robustness-scope}: rare-feature sensitivity across token budgets, the movement a boundary intervention buys against what it costs, and the same comparison read within each model family.

\subsection{A boundary view of feature sensitivity}
\label{sec:boundary-stability-view}

The margin diagnosis can be written as a selection-boundary statement. Any sparse SAE must decide whether feature $i$ is available on activation $z$. Let $c_i(z)$ denote the boundary that the feature's score $a_i(z)$ must exceed, and define the feature margin and the active indicator
\begin{equation}
  m_i(z)=a_i(z)-c_i(z), \qquad A_i(z)=\mathbf{1}\{m_i(z)>0\},
\end{equation}
so that, together with the post-ReLU sign of Section~\ref{sec:preliminaries}, a feature is active exactly when $m_i(z)>0$. For a TopK SAE this boundary is a rank boundary. Let $a_{-i,(k)}(z)$ be the $k$th largest score among all features except $i$; then feature $i$ is in the TopK active set exactly when
\begin{equation}
  m_i^{\mathrm{TopK}}(z)=a_i(z)-a_{-i,(k)}(z)>0.
\end{equation}
For a source-active feature, $a_{-i,(k)}(z)$ is the first rejected competitor, so this recovers the active-margin diagnostic used in the main text, and the example-level top-$k$ margin $a_{(k)}(z)-a_{(k+1)}(z)$ is its global summary of how crowded the cutoff is. Pairwise rank stabilization replaces the single rejected competitor with a smooth centered boundary over a band of hard negatives,
\begin{equation}
  B_R(z) = \tau\left(\log\sum_{j\in \mathcal{H}_R(z)}\exp(a_j(z)/\tau)-\log R\right),
  \qquad
  m_i^{R}(z)=a_i(z)-B_R(z),
\end{equation}
where $\mathcal{H}_R(z)$ contains the $R$ features just below the cutoff. This is the same TopK boundary idea written at the level of a single feature, and the loss that acts on it is the ordering objective of Algorithm~\ref{alg:rank-stability-sae}.

Under a meaning-preserving perturbation such as a paraphrase, write the paired activation as $z'$ and define the margin change
\begin{equation}
  m_i(z') = m_i(z)+\eta_i,
  \qquad
  \eta_i = m_i(z')-m_i(z).
\end{equation}
This decomposition is exact once $\eta_i$ is defined as the paraphrase-induced change in distance to the selection boundary, so a source-active feature drops exactly when that change crosses the negative source margin. If $F_{i,z}$ denotes the local cumulative distribution function of $\eta_i$ under paraphrases of $z$, then
\begin{equation}
  \Pr[\text{drop}_i\mid z]
  = \Pr[m_i(z')<0]
  = \Pr[\eta_i<-m_i(z)]
  = F_{i,z}(-m_i(z)),
\end{equation}
which gives the boundary prediction before any parametric form is imposed: when the local perturbation law is comparable across features, smaller source margins imply a larger paraphrase drop probability. A Gaussian local approximation, $\eta_i\mid z \sim \mathcal{N}(\mu_{i,z},\sigma_{i,z}^2)$, then gives the closed form
\begin{equation}
  \Pr[\text{drop}_i\mid z]
  = \Phi\left(\frac{-m_i(z)-\mu_{i,z}}{\sigma_{i,z}}\right),
\end{equation}
where $\Phi$ is the standard normal cumulative distribution function. Here $\mu_{i,z}$ and $\sigma_{i,z}$ summarize the local drift and scale of paraphrase-induced margin changes; they are not additional SAE parameters and do not require independence across features. This yields the monotone prediction tested in Figure~\ref{fig:width-margin}c: source-active features closer to the cutoff should drop more often under paraphrase. Rare features are not assumed to be semantically special by definition; they become the reliability stress test because, in the studied TopK scaling setting, many of their active instances sit closer to the selection boundary.

Other sparse mechanisms define different availability boundaries. JumpReLU keeps a feature only when its score exceeds a learned feature-specific threshold~\citep{rajamanoharan2024jumprelu}, giving $m_i^{\mathrm{thr}}(z)=a_i(z)-\theta_i$. BatchTopK uses a rank boundary across a batch during training and is often evaluated with an estimated global threshold at inference~\citep{bussmann2024batchtopk}. Gated SAEs and sparse routing mechanisms, including mixture-of-experts style routers, impose analogous availability boundaries. The relevant margin therefore depends on the sparsity mechanism: TopK repairs stabilize rank ordering around a cutoff, whereas threshold repairs would stabilize distance to a learned threshold.

We use a public JumpReLU SAE from Gemma Scope to check that the same boundary intuition appears under a learned threshold. The check uses \texttt{google/gemma-2-9b} with Gemma Scope \texttt{layer\_14/width\_16k/average\_l0\_67}, 148 evaluation pairs, and 9174 source-active feature instances. Table~\ref{tab:jumprelu-boundary-diagnostic} reports the threshold-margin trend. Over all instances, source margin is negatively correlated with drop probability (Spearman $\rho=-0.4573$, logistic coefficient $-0.8838$), consistent with the boundary diagnosis beyond TopK and leaving threshold repair as a separate design question.

The boundary account is not specific to TopK. On a public JumpReLU SAE, whose features carry learned thresholds, the same monotone pattern appears: the drop rate falls from $61.8\%$ in the nearest-threshold bin to $1.5\%$ in the farthest (Table~\ref{tab:jumprelu-boundary-diagnostic}).

\begin{table}[!htbp]
\centering
\caption{\textbf{JumpReLU threshold-margin diagnostic.} In a JumpReLU SAE each feature has a learned threshold, so the margin is $a_i(z)-\theta_i$. Rows bin 9174 source-active instances from one public Gemma Scope checkpoint by source-side threshold margin, nearest first; \emph{Drop} and \emph{Retain} are percentages.}
\label{tab:jumprelu-boundary-diagnostic}
\fittable{%
\begin{tabular}{@{}llrrrr@{}}
\toprule
\bfseries
Bin & \bfseries Source-margin range & \bfseries $n$ & \bfseries Drop $\downarrow$ & \bfseries Retain $\uparrow$ & \bfseries Mean source / paraphrase margin \\
\midrule
\normalfont
near-threshold & $[0.0005,0.196)$ & 918 & 61.76 & 38.24 & 0.096 / -0.338 \\
low & $[0.196,0.546)$ & 1376 & 48.47 & 51.53 & 0.365 / -0.019 \\
mid-low & $[0.546,1.411)$ & 2292 & 28.58 & 71.42 & 0.935 / 0.544 \\
mid-high & $[1.411,3.131)$ & 2294 & 11.29 & 88.71 & 2.131 / 1.678 \\
high & $[3.131,7.584)$ & 1376 & 3.78 & 96.22 & 4.696 / 3.954 \\
far-from-threshold & $[7.584,141.850]$ & 918 & 1.53 & 98.47 & 16.463 / 15.496 \\
\bottomrule
\end{tabular}}
\end{table}

\FloatBarrier

\subsection{Additional Controls}
Table~\ref{tab:fullscale-margin-flip} links active margin to actual paraphrase drops. A specificity audit finds that boundary manipulation does not simply turn rare features into broadly active ones: migration from rare to common and false positives on unrelated text both stay low.

\paragraph{Loss-variant definitions.}
The additional loss controls use the same notation as the margin diagnosis. Let $\delta$ be the target clearance and let
$B_R(z)=\tau\{\log\sum_{j\in\mathcal{H}_R(z)}\exp(a_j(z)/\tau)-\log R\}$ denote a smooth hard-negative boundary over the $R$ scores below the TopK cutoff. The hinge boundary control penalizes $[\delta-(a_{(k)}(z)-a_{(k+1)}(z))]_+$, directly enlarging the global top-$k$ gap. The LSE boundary control replaces the single $(k+1)$st competitor with $B_R(z)$, giving $[\delta-(a_{(k)}(z)-B_R(z))]_+$. The rare-band margin control applies the same clearance penalty only to rare selected features, averaging $[\delta-(a_i(z)-B_R(z))]_+$ over rare $i\in\operatorname{TopK}(a(z))$. Rank-stability denotes the pairwise ordering loss in Algorithm~\ref{alg:rank-stability-sae}: it compares source-active rare features with paraphrase hard competitors, targeting stability of the ordering rather than merely increasing the global margin. Table~\ref{tab:main-core-result} compares these controls at the 10M token budget.

\FloatBarrier

\subsection{SAE-internal usability proxies}
\label{app:usability-proxies}

Table~\ref{tab:feature-usability-checks} reports two SAE-internal summaries of the active set for rare features. Both describe how the active set itself behaves under paraphrase rather than any downstream task: true retention is the fraction of source-active cases retained on the true paraphrase, mismatch activation uses a paraphrase from another source as a negative, and effect retention is the paraphrase/source ratio of the SAE feature contribution mass for source-active cases.

\begin{table}[H]
\centering
\caption{\textbf{SAE-internal usability proxies for rare features.} Both columns are internal summaries of the active set, not downstream task metrics. True retention is the fraction of source-active cases retained on the true paraphrase; mismatch activation uses a paraphrase from another source as a negative, and retention gap is their difference. Effect retention is the paraphrase/source ratio of the SAE feature contribution mass for source-active cases. Values are mean$\pm$SEM over seeds, with $n$ the number of seeds.}
\label{tab:feature-usability-checks}
\begingroup
\setlength{\tabcolsep}{3.0pt}
\renewcommand{\arraystretch}{1.05}
\fittable{%
\begin{tabular}{@{}lccccc@{}}
\toprule
\bfseries
Method & \bfseries $n$ & \bfseries True ret. $\uparrow$ & \bfseries Mismatch act. $\downarrow$ & \bfseries Retention gap $\uparrow$ & \bfseries Effect ret. $\uparrow$ \\
\midrule
\normalfont
baseline & 5 & 0.690$\pm$0.019 & 0.452$\pm$0.028 & 0.238$\pm$0.011 & 0.735$\pm$0.019 \\
rank-stability & 5 & \textbf{0.755$\pm$0.008} & 0.507$\pm$0.017 & \textbf{0.248$\pm$0.011} & \textbf{0.791$\pm$0.011} \\
\bottomrule
\end{tabular}}
\endgroup
\end{table}

\FloatBarrier

\section*{AI Use Disclosure}

Language models were used for drafting, restructuring, and language editing, and a second, independently prompted language model reviewed the manuscript for internal consistency; the authors verified all technical claims, numbers, and citations. Language models were \emph{not} used as annotators or as components of any method evaluated in this paper. The one exception is that the paraphrase pairs used as a reliability probe were generated by GPT-4o under the prompt described in the appendix; each generated pair was manually reviewed for meaning preservation before use, and no model output was used as a label or as ground truth.

\end{document}